\documentclass[manuscript,nonacm,screen]{acmart}

\AtBeginDocument{%
  }

\usepackage{color}
\usepackage{graphicx,pifont}
\usepackage{subcaption}
\usepackage{enumitem}
\usepackage{xspace}
\usepackage{pifont}
\author{Hanlin Du}
\affiliation{
  \institution{SKLP, Institute of Computing Technology, CAS}
  \city{Beijing}
  \country{China}
}
\affiliation{
  \institution{University of Chinese Academy of Sciences}
  \city{Beijing}
  \country{China}
}
\email{duhanlin21@mails.ucas.ac.cn}

\author{Zhiyuan Yan}
\affiliation{
  \institution{SKLP, Institute of Computing Technology, CAS}
  \city{Beijing}
  \country{China}
}
\email{yanzhiyuan@ict.ac.cn}

\author{Yungang Bao}
\affiliation{
  \institution{SKLP, Institute of Computing Technology, CAS}
  \city{Beijing}
  \country{China}
}
\affiliation{
  \institution{University of Chinese Academy of Sciences}
  \city{Beijing}
  \country{China}
}
\email{baoyg@ict.ac.cn}

\author{Sa Wang}
\affiliation{
  \institution{SKLP, Institute of Computing Technology, CAS}
  \city{Beijing}
  \country{China}
}
\affiliation{
  \institution{University of Chinese Academy of Sciences}
  \city{Beijing}
  \country{China}
}
\email{wangsa@ict.ac.cn}

\newcommand{\workname}{\textsc{DynaResize} }
\newcommand{\para}[1]{\smallskip\noindent\textbf{#1}}
\begin{document}

\title{\workname: Runtime GPU Reallocation for Disaggregated LLM Post-Training}


\begin{abstract}
RL-based LLM post-training increasingly disaggregates Rollout and Training across separate GPU resources, but static GPU partitioning suffers from severe pipeline bubbles under long-tail rollout latency. We present \workname, a runtime GPU reallocation system that dynamically switches GPUs between Rollout and Training to balance stage execution times without changing RL semantics. \workname decomposes resizing into fine-grained operations and removes non-startup-critical work from the critical path through communicator reuse, bounded state staging, and hysteresis-based resizing. Experimental results show that \workname can improve end-to-end throughput by 66.5\% and reduce total execution time by 33\% over the optimal static configuration, while hiding 27\% of role-switching overhead.
\end{abstract}

\maketitle

\section{Introduction}

Large Language Model (LLM) post-training has become a critical phase in modern AI infrastructure \cite{cite:qwen3, cite:deepseekv4,cite:llama3}. Unlike standard pre-training, reinforcement learning (RL) based post-training follows an iterative two-stage workflow, where \textit{Rollout} generates trajectories using the current policy and \textit{Training} updates model parameters based on the collected experience  \cite{cite:instructgpt, cite:ppo, cite:grpo, cite:dapo, cite:dpo}. A straightforward sequential execution of these two stages leads to poor pipeline efficiency, as rollout and training cannot make progress concurrently. To address this limitation, modern distributed frameworks such as veRL \cite{cite:verl-paper, cite:verl}, slime \cite{cite:slime}, RLinf \cite{cite:rlinf-paper, cite:rlinf}, and AReaL \cite{cite:areal-paper, cite:areal} decouple Rollout and Training through disaggregated deployment and coordinated scheduling, enabling the two stages to overlap across different workers and thereby improving concurrency and end-to-end performance. 

However, this disaggregated paradigm remains fundamentally vulnerable to unpredictable Rollout latencies \cite{cite:vllm, cite:orca, cite:multibin}. Since generation length varies significantly with prompt complexity, Rollout often exhibits severe long-tail behavior, leaving downstream Training stalled and degrading end-to-end system efficiency. Recent efforts seek to hide this inefficiency through \textit{algorithmic} compromises, such as truncating rollout lengths \cite{cite:kimi,cite:partial-rollout,cite:tppo} or tolerating stale policy updates \cite{cite:asyncflow}. Yet these workarounds come at the cost of deviating from the original reinforcement learning semantics, potentially compromising the final model accuracy.

In this paper, we ask the question: \textit{Can long-tail bubbles in disaggregated post-training be mitigated through a system-level solution?} Specifically, we explore whether GPUs can be dynamically reassigned between Rollout and Training at runtime, allowing resources to follow the shifting bottleneck and reduce pipeline stalls caused by long-tail Rollout latencies. However, making this idea practical is far from straightforward. Dynamic resizing may incur substantial overheads from worker reconfiguration, state migration, and runtime coordination, potentially diminishing or even outweighing its performance benefits. Moreover, existing distributed frameworks provide no direct support for dynamic GPU reallocation between Rollout and Training. If implemented naively, such reallocation would introduce three fundamental challenges.

\ding{182} \textit{Topology Reconstruction.} Switching the role of a GPU requires destroying and rebuilding the global communication topology. This monolithic barrier synchronization across nodes incurs pipeline stalls lasting minutes.
\ding{183} \textit{Parameter Migration Overflow.} Dynamic reconfiguration requires the target GPU to flush old parameters and ingest new weights and optimizer states. Under tight GPU memory constraints, this in-place context switch may cause severe transient memory amplification, triggering Out-of-Memory (OOM) crashes.
\ding{184} \textit{Scheduling Thrashing.} LLM generation length exhibits extreme token-level variance. If the system reacts passively to instantaneous load imbalances, it suffers from severe \textit{thrashing}. In this state, GPUs spend more time on context-switching than on useful computation. This overhead completely negates the theoretical benefits of dynamic allocation.
Consequently, prior systems are forced to settle for the static paradigm. Overcoming these systemic barriers demands a fundamental redesign of the resource scaling mechanism.

To this end, we propose \workname, a dynamic resource-scaling system that enables low-overhead GPU role switching between Rollout and Training on the fly for disaggregated post-training. The key idea of \workname is to revisit the resource resizing process by decomposing it into fine-grained operations and analyzing their execution dependencies. Rather than performing resizing as a monolithic stop-and-wait procedure, \workname determines which operations can run in parallel, which can be prewarmed ahead of topology activation, and which can be deferred until they are actually needed. It then moves expensive but non-startup-critical operations, such as communicator preparation and optimizer-state reloading, off the critical path. In this way, dynamic GPU resizing becomes a lightweight runtime operation instead of a disruptive global reconfiguration.

Building on this core idea, \workname further implements three mechanisms to address the key challenges of dynamic GPU reallocation: \textit{communicator reuse}, \textit{state staging}, and \textit{hysteresis resizing}. Communicator reuse reduces topology switching overhead by caching topology-scoped communication metadata and prewarming inter-role communicators off the critical path. State staging minimizes state migration latency by streaming model weights and optimizer states through bounded host-memory buffers, while deferring non-critical optimizer-state reloading until it is actually needed. Hysteresis resizing stabilizes GPU reallocation decisions by triggering resizing only under sustained workload imbalance rather than transient latency fluctuations.

Crucially, \workname does not reduce individual rollout latency. Instead, it improves cluster-wide utilization by rebalancing GPUs between Rollout and Training at runtime. It is also orthogonal to prior algorithmic methods and can be composed with them for extreme long-tail workloads.

Evaluations demonstrate that compared to the optimal static configuration, \workname promptly rectifies systemic imbalances at runtime. By fluidly scaling resource partitions, \workname improves end-to-end post-training throughput by 66.5\% and shortens the total execution time by 33\% (normalized to a 100-step baseline), while reducing resizing overhead by 27\% compared to rigid refactoring.

\section{Background and Motivation}
\label{sec:bg&moti}

\begin{figure}[t]
    \centering
    \includegraphics[width=0.9\linewidth]{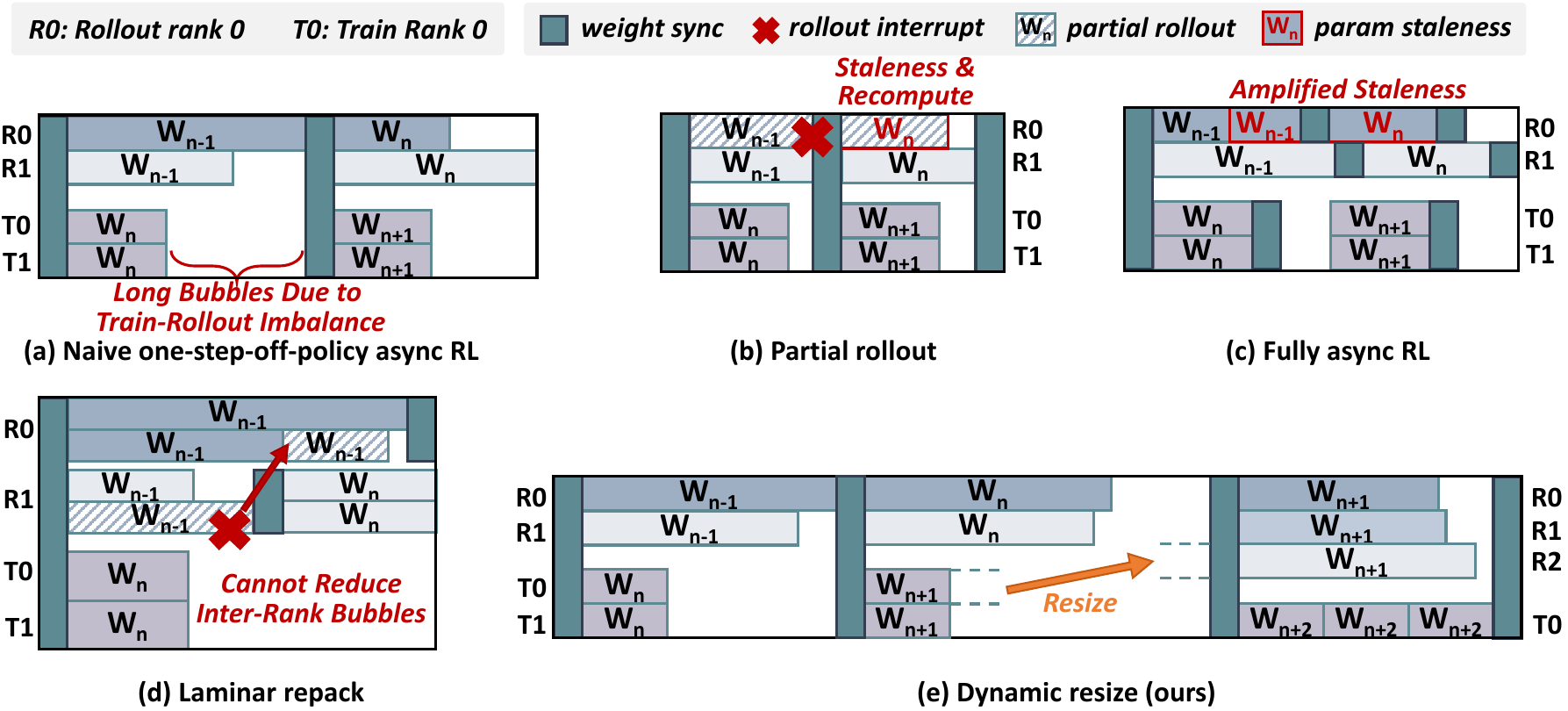}
    \caption{Comparison of approaches for mitigating asynchronous training-rollout imbalance}
    \Description{Comparison of one-step-off-policy and its 3 algo solutions, and our work}
    \label{fig:relwork}
\end{figure}

Modern LLM post-training relies on asynchronous pipelines to physically decouple the rollout stage from the training stage. While this separation mitigates strict synchronization overhead, it exposes a critical flaw under static resource provisioning: the highly variable generation lengths lead to severe workload imbalances. As illustrated in Figure \ref{fig:relwork}(a), under a naive one-step-off-policy asynchronous execution, rollout stragglers induce massive pipeline bubbles, leaving allocated training GPUs idling. Furthermore, the execution time of rollout steps often significantly exceeds that of training steps (especially for complex agentic or reasoning tasks), exacerbating these idle periods.

To alleviate this hardware under-utilization, recent algorithmic and scheduling modifications navigate various trade-offs (Figure \ref{fig:relwork}(b)-(d)). 
\begin{itemize}[leftmargin=*, topsep=1pt]
\item \textbf{Partial Rollout \& Truncation. (Fig. \ref{fig:relwork}(b))} Approaches like Kimi~\cite{cite:kimi} and TPPO~\cite{cite:tppo} enforce time alignment by interrupting rollouts, but this compromises the integrity of long-form reasoning (e.g., Chain-of-Thought). 
\item \textbf{Fully Asynchronous RL. (Fig. \ref{fig:relwork}(c))} Frameworks such as AsyncFlow \cite{cite:asyncflow} maximize hardware overlap by decoupling stages entirely; however, this inherently introduces severe policy staleness, which risks convergence and requires careful algorithmic compensation. 
\item \textbf{Advanced Repacking. (Fig. \ref{fig:relwork}(d))} Strategies like Laminar~\cite{cite:laminar} attempt to mask latency by tightly packing requests, yet they still struggle to fully eliminate inter-rank bubbles when intra-batch variance is extreme.
\end{itemize}





%
Moreover, methods attempting to mitigate variance by rearranging samples via curriculum-learning-based length prediction suffer from poor prediction accuracy and degrade the model's long-context capabilities~\cite{cite:sortedrl,cite:rollpacker}. 
From an infrastructure perspective, recent elastic frameworks scale rollout ranks by harvesting external GPUs~\cite{cite:rose,cite:echo2,cite:prorl}. Yet, relying on extra hardware solely for the rollout phase lacks the system-level control to balance the train-rollout loop, and fails to eliminate idle waste during pipeline stalls as well \cite{cite:torchelastic,cite:optimus,cite:bamboo,cite:oobleck}.

%

We argue that pipeline imbalance is primarily a physical resource mismatch rather than an algorithmic flaw. Therefore, a natural system-level approach is to dynamically reallocate resources between Training and Rollout, as shown in Figure~\ref{fig:relwork}(e). In particular, addressing this imbalance requires an intra-cluster closed-loop mechanism that can convert existing GPUs between the two stages on demand, reducing hardware idling during pipeline stalls while preserving the original algorithmic semantics.

\section{\textsc{DynaResize}'s Design}
\label{sec:design}


\begin{figure}[t]
    \centering
    \includegraphics[width=0.8\linewidth]{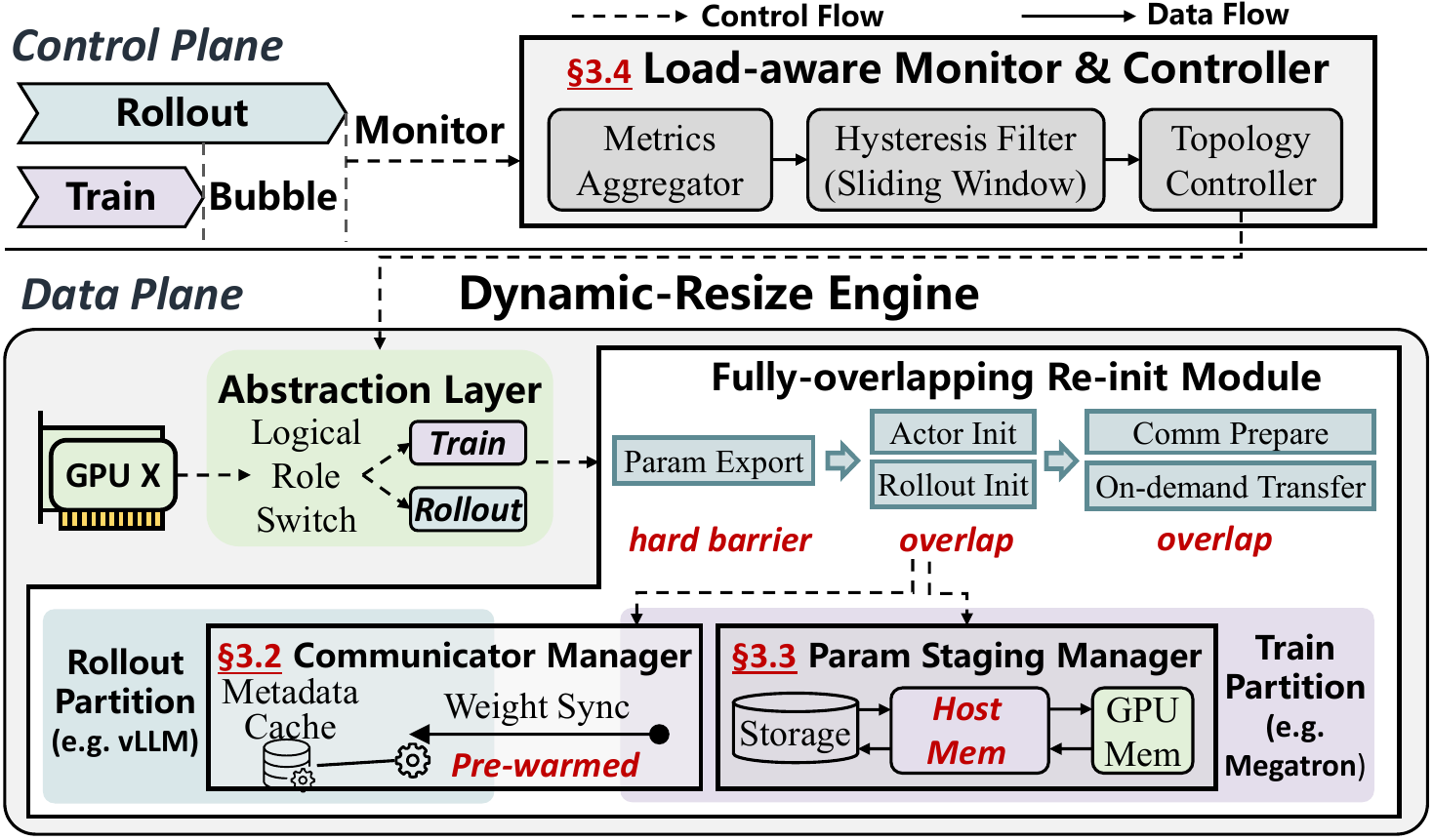}
    \caption{Overview of \workname}
    \Description{This fig shows the design of \workname, which contains scheduler, abstraction layer, re-init module, comm \& data store manager}
    \label{fig:arch}
\end{figure}

\subsection{Overview}

As illustrated in Fig.\ref{fig:arch}, \workname adopts a decoupled control-data plane architecture centered around an abstraction layer that isolates physical GPUs from logical roles, allowing on-the-fly computational reallocation. On the control plane, a load-aware monitor and a topology controller work in tandem to track pipeline bubbles and conservatively issue resize commands, which is detailed in \S\ref{subsec:schedule}.

\begin{figure}[t]
    \centering
    \includegraphics[width=0.9\linewidth]{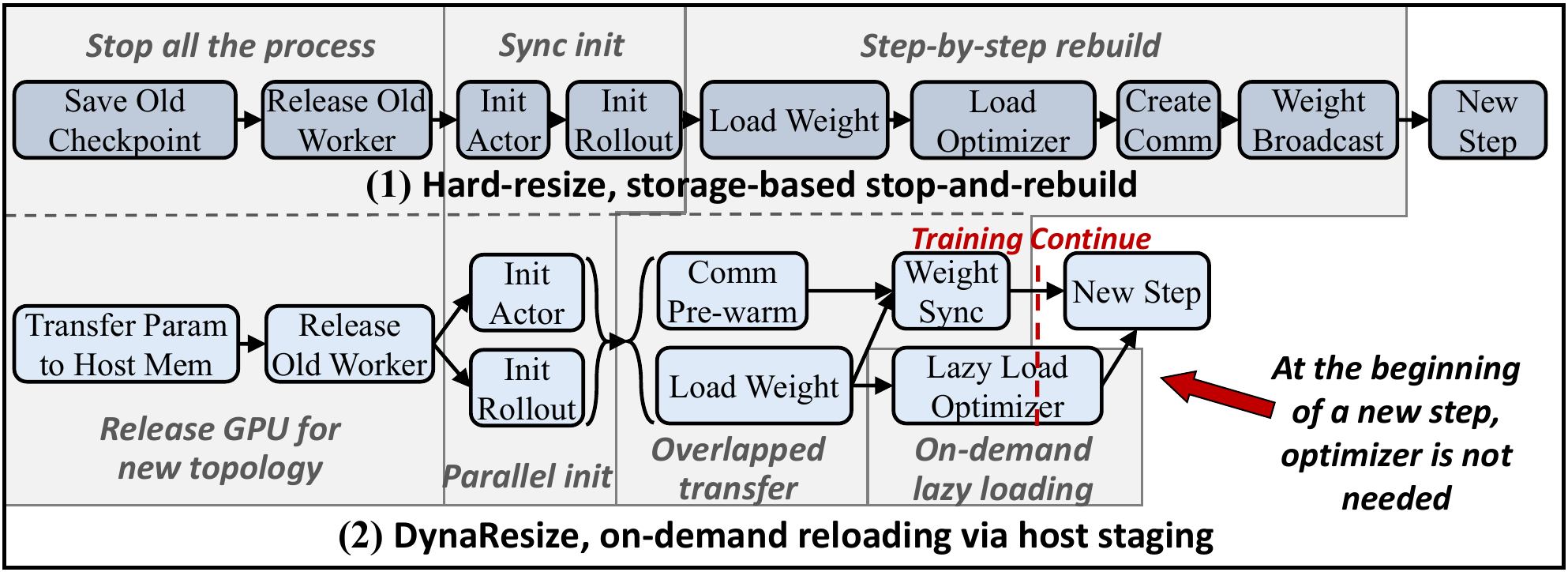}
    \caption{Fully-overlapping workflow of \workname compared to hard stop-and-rebuild resizing}
    \Description{workflow of \workname and hard resize, showing overlapping}
    \label{fig:workflow}
\end{figure}

Fig. \ref{fig:workflow} highlights the resulting paradigm shift by contrasting the execution timelines of a traditional hard-resize against our non-blocking \workname workflow. To fundamentally resolve the prohibitive overhead of the resizing process, \workname abandons the monolithic, \textit{stop-and-rebuild} lifecycle. Instead, it pipelines role-switching under the centralized orchestration of the \textbf{Fully-overlapping Re-init Module}. This manager concurrently overlaps most parallelizable operations not logically constrained and deferred parameter transfers with the ongoing computation. To achieve this massive overlapping, it governs two subordinate data-plane strategies: \textbf{(1) Topology Reconstruction (\S\ref{subsec:comm}):} The Communicator Manager pre-warms communication metadata off the critical path to reduce rebuilding barriers. \textbf{(2) Parameter Migration (\S\ref{subsec:host_staging}):} The Param Staging Manager safely pipelines weights and optimizer states via a chunked streaming mechanism to prevent memory crashes.



\subsection{Topology-Aware Communicator Reuse}
\label{subsec:comm}

\textbf{Problem.} Communication reconstruction is tightly coupled with the worker lifecycle. In existing distributed runtimes, communication groups cannot be fully established until the corresponding workers have been initialized and their ranks, devices, and runtime contexts become available \cite{cite:pytorch,cite:nccl}. As a result, each topology switch may require rebuilding inter-role communication groups on the critical path before weight synchronization can proceed. This creates a rigid reconfiguration overhead. Although the cost cannot be completely eliminated due to the dependency on worker initialization, leaving it entirely on the switching path would significantly delay GPU role transitions.

\textbf{Solution.} \workname mitigates this overhead by separating topology-scoped communication metadata from worker-bound runtime states. For each candidate topology, \workname caches the metadata required to construct inter-role communication groups, such as rank mappings and group membership. Once the corresponding workers become available, \workname prewarms the required communicators off the critical path before the new topology is published. In this way, topology switching no longer rebuilds communication groups from scratch. Instead, the actual switch becomes a lightweight communicator binding and activation step, which accelerates subsequent weight synchronization.

Crucially, our communicator reuse is conservative by design. It applies specifically to topology-scoped inter-role synchronization groups for model weights. We do not claim full reuse of worker-bound distributed runtime states (such as the default process group or device-mesh bindings), as complete runtime communication reuse would require a persistent refactoring of workers, violating the modular structure of existing frameworks. This approach recovers substantial setup costs across revisited topologies while preserving framework correctness.

\subsection{Staged State Reloading}
\label{subsec:host_staging}

\textbf{Problem.} Elastic reconfiguration requires migrating the training state across different GPU topologies. During resizing, training ranks must preserve the current model weights and optimizer states, and then reload the states required by the new topology \cite{cite:megatron,cite:fsdp}. A naive implementation would follow a storage-based checkpointing path: it first flushes the existing device states to external storage, releases the old workers, and then reloads the states for the new workers. This stop-and-reload procedure introduces a long blocking period at the step boundary. Since optimizer states are often several times larger than model weights, full-state reconstruction can severely stall subsequent training steps.

\textbf{Solution.} \workname avoids this storage-based path by using host memory as a fast staging area for state migration. \workname utilizes CPU-offloading method \cite{cite:zero,cite:zero-offload,cite:zero-infinity,cite:flexgen}, directly transferring weights and optimizer states from old training ranks to new training ranks through host memory. However, naively staging the entire state in host memory can easily cause out-of-memory failures. To keep memory usage bounded, \workname partitions the states into fine-grained chunks and streams them through host-memory buffers. The store and reload operations are then pipelined chunk by chunk. This design avoids full-state accumulation in host memory, while still preserving the speed advantage of in-memory transfer over storage-based checkpointing.

Crucially, the core efficiency of \workname gains originate from maximizing the asynchronous overlapping of independent operations and enforcing an on-demand optimizer reload. Considering that optimizer states are several times larger than model weights and are only required at the end of a training step, we design an on-demand deferred transfer mechanism. Only the model weights are loaded during the immediate transition phase to resume step execution, while the loading and materialization of the massive optimizer states are deferred until the first step on the new topology. This strategic deferral successfully absorbs the heaviest data-transfer overhead from the critical path of the step boundary, enabling low-overhead topology transitions under strict resource constraints.

\subsection{Hysteresis-Guided Reconfiguration}
\label{subsec:schedule}
\textbf{Problem.} Resizing decisions are vulnerable to short-term workload noise. In RL post-training, Rollout and Training costs can vary significantly across batches, causing transient throughput fluctuations. A naive controller that reacts to every measurement may frequently switch GPU allocations back and forth. Such reconfiguration thrashing can introduce repeated resizing overheads and may even outweigh the load-balancing benefits of dynamic reallocation.

\textbf{Solution.} \workname treats resizing as a stabilized decision problem over a discrete set of candidate topologies. Instead of using each measurement as an immediate trigger, \workname filters resizing decisions through hysteresis thresholds, dwell-time constraints, cooldown intervals, and consecutive-signal requirements. As a result, resizing is triggered only by sustained directional imbalance rather than transient fluctuations. The controller does not perform unconstrained online topology search. It only decides when to transition among pre-specified elastic configurations, ensuring that each resizing action can bring positive throughput gains.

\section{Implementation}
\label{sec:implementation}

We implement \workname atop veRL \cite{cite:verl}, a state-of-the-art LLM post-training framework, which uses Ray \cite{cite:ray} to statically provision GPU actor groups for training and rollout. We embed our \workname engine directly into this worker lifecycle. Instead of triggering Ray's heavy actor teardown during a resize, \workname retains the underlying placement groups. By leveraging host-memory staging and communicator caching, \workname dynamically migrates model states and transitions execution roles within existing processes, enabling heavyweight resizing operations to be executed asynchronously and fully overlapped.

\section{Evaluation}
\label{sec:evaluation}

Our evaluation is designed to answer three questions regarding the performance and overhead of \workname.
\begin{itemize}[leftmargin=*, topsep=1pt]
    \item How much performance improvement can \workname achieve compared to optimal static configurations? \textbf{(Exp\#1)}
    \item How much reconfiguration latency can \workname hide compared to a naive \textit{stop-and-rebuild} baseline? \textbf{(Exp\#2)}
    \item How many training steps are required to amortize the cost of a single resize operation, and does the hysteresis controller successfully prevent thrashing? \textbf{(Exp\#3)}
\end{itemize}

\begin{figure}[t]
    \centering
    \begin{subfigure}{0.45\linewidth}
        \includegraphics[width=\linewidth]{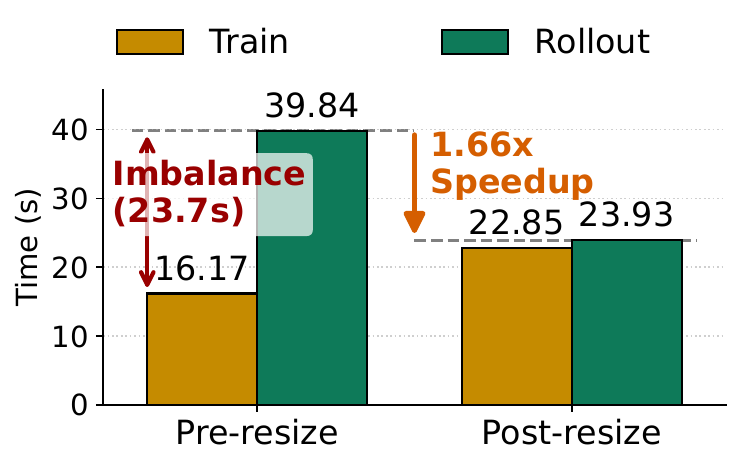}
        \caption{Avg step time.}
        \label{fig:steptime_left}
    \end{subfigure}
    \begin{subfigure}{0.45\linewidth}
        \includegraphics[width=\linewidth]{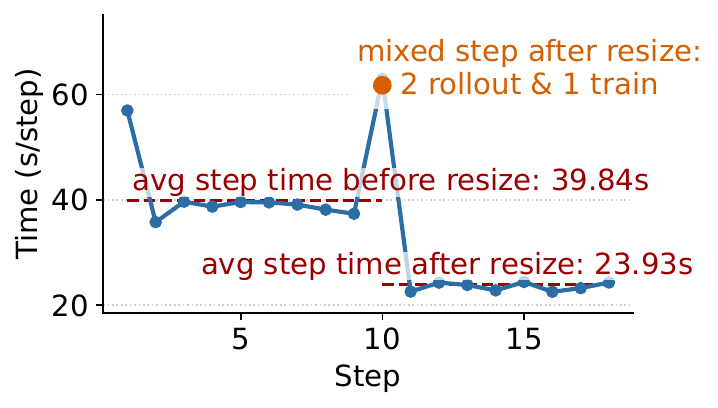}
        \caption{Time per step.}
        \label{fig:steptime_right}
    \end{subfigure}
    \caption{\textbf{(Exp\#1)} Time per step saved by dynamic resizing.}
    \Description{Exp for throughput speedup, about 1.66x}
    \label{fig:steptime}
\end{figure}
\begin{figure*}[t]
    \centering
    \includegraphics[width=\linewidth]{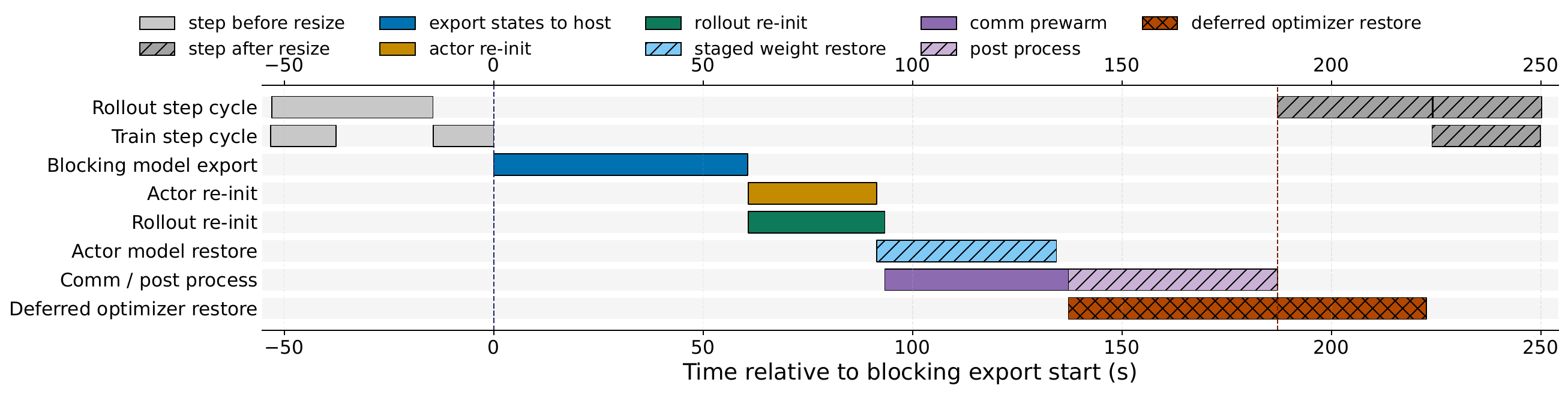}
    \caption{\textbf{(Exp\#2)} Time breakdown for \workname comparing to hard resize without on-demand reloading.}
    \Description{Exp that profiles resize cost}
    \label{fig:time_flame}
\end{figure*}

\begin{figure}[t]
    \centering
    \includegraphics[width=0.8\linewidth]{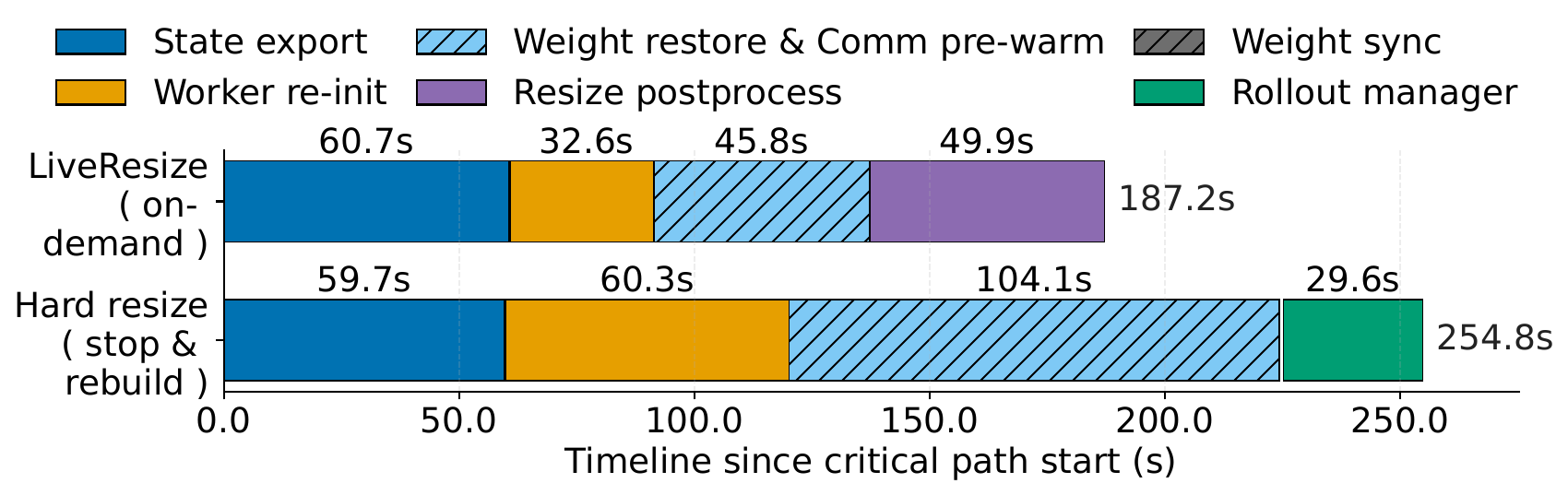}
    \caption{\textbf{(Exp\#2)} Time flow of \workname caching and overlapping.}
    \Description{Exp compares breakdown cost of \workname with hard resize}
    \label{fig:breakdown}
\end{figure}

\para{Experimental Setup.} We evaluate \workname on a GPU server with 8$\times$ NVIDIA H20 GPUs connected via NVLink and PCIe Gen5. We use Qwen3-8B as the training model. To simulate real-world post-training imbalances, we utilize a variance prompt dataset by mixing short-form QA tasks with long-form reasoning and coding tasks, deliberately inducing severe generation long-tail stragglers.

\para{(Exp\#1)~End-to-End Performance.}
We first evaluate end-to-end performance under different degrees of workload variance, measured by the execution time per PPO step. Under high prompt variance, \workname improves end-to-end throughput by 66.5\% and reduces the total execution time by 33\% compared with the optimal static baseline, as shown in Figure~\ref{fig:steptime}(a). These results are reported over a normalized 100-step run, while practical post-training jobs typically execute 500--5000 steps. This performance gain is primarily achieved by aligning the execution times of Rollout and Training within each PPO step. \workname mitigates imbalance by dynamically reallocating GPUs at runtime (e.g., shifting the Train-to-Rollout ratio from 6:2 to 4:4 to match the current bottleneck). As detailed in Figure~\ref{fig:steptime}(b), at the point of reallocation (Step 10), the step time temporarily spikes. This prolonged execution is caused by the overhead of a "mixed step" during resizing (e.g., executing 2 rollouts and 1 train). However, immediately after this resize step, the system stabilizes into an ideal state with the average step time drastically reduced (from 39.84s down to 23.93s). This effectively eliminates Rollout-induced pipeline bubbles and maximizes global GPU utilization.


\textbf{Answer\#1:} \textit{\workname improves throughput by 66.5\% by aligning Rollout and Training times through runtime GPU reallocation.}



\para{(Exp\#2)~Resizing Breakdown.}
We next analyze the latency breakdown of a single GPU role transition, such as switching GPUs from Training to Rollout. We compare \workname with a naive \textit{stop-and-rebuild} resizing baseline, which performs role switching without our critical-path optimizations.

Figures~\ref{fig:time_flame} and~\ref{fig:breakdown} show the detailed latency breakdown of the reconfiguration phase. Under the naive baseline, each role switch incurs a monolithic blocking period of 255 seconds. This delay is mainly caused by Ray actor destruction, device-state flushing, and the blocking execution of \texttt{dist.init\_process\_group} for rebuilding the communication topology. By overlapping in parallel as much as possible, \workname reduces this blocking cost to 187 seconds by moving non-startup-critical operations off the critical path. 

\textbf{Answer\#2:}~\textit{Although role switching still involves unavoidable synchronous operations, \workname reduces the critical-path blocking time from 255 seconds to 187 seconds, hiding 27\% of the overhead through background execution.}

\para{(Exp\#3)~Cost Amortization and Thrashing Boundaries.}
We further analyze whether the benefit of \workname can reliably outweigh its reconfiguration cost, and whether the resizing policy remains stable under transient workload noise. 

\underline{\textit{Cost amortization.}} \workname becomes beneficial once the saved pipeline idle time offsets the critical-path switching latency. Our evaluation identifies a break-even point of approximately 11 sustained steps. This brief amortization window ensures positive throughput gains in following steady steps whenever train-rollout imbalances persist beyond a short period.


\underline{\textit{Thrashing boundaries.}} Unconstrained resizing risks reacting to short-term fluctuations, causing reconfiguration thrashing that negates load-balancing benefits. To prevent this, our Hysteresis-Guided controller enforces a >10-step cooldown interval, safely exceeding the amortization threshold. This efficiently filters high-frequency noise, triggering migrations exclusively during sustained directional imbalances to guarantee strictly positive gains

\textbf{Answer\#3:}~\textit{\workname amortizes the cost of a single resize operation within $\sim$11 sustained training steps. The hysteresis controller can prevent thrashing.}




\section{Conclusion}
\label{sec:conclusion}


\workname enables runtime dynamic GPU resizing for RL-based LLM post-training, reallocating GPUs between Rollout and Training to mitigate pipeline bubbles without algorithmic approximations. Although pre-warming and on-demand loading only moderately reduce the critical-path resizing latency, this reduction is sufficient to yield substantial end-to-end gains in pipeline-stall-dominated post-training workloads. In our evaluation, hiding 27\% of role-switching overhead enables \workname to improve end-to-end throughput by 66.5\% over the optimal static configuration. These results suggest that runtime resource reallocation is a promising system-level direction for improving post-training efficiency, while preserving RL semantics and algorithmic effectiveness.

\newpage

\bibliographystyle{ACM-Reference-Format}
\bibliography{ref}




\end{document}